\documentclass{svproc}
\usepackage{url}
\usepackage{graphicx}

\begin{document}
\mainmatter              
\title{Towards sample-efficient policy learning with DAC-ML}
\titlerunning{DAC-ML}  
%
\author{Ismael T. Freire\inst{1,2,3} \and Adrián F. Amil\inst{1,2,3}
 \and Vasiliki Vouloutsi\inst{1,3} \and \\ Paul F.M.J. Verschure \inst{1,2,3,4}
\thanks{ITF and AFA contributed equally to this work}
}
\authorrunning{Ismael T. Freire et al.} 
%

%
\institute{Institute for Bioengineering of Catalonia (IBEC), Barcelona, Spain \\
\email{ifreire@ibecbarcelona.eu}
\and Universitat Pompeu Fabra (UPF), Barcelona, Spain 
\and Barcelona Institue of Science and Technology (BIST), Barcelona, Spain 
\and Catalan Institute for Research and Advanced Studies (ICREA), Barcelona, Spain}

\maketitle              

\begin{abstract}
The sample-inefficiency problem in Artificial Intelligence refers to the inability of current Deep Reinforcement Learning models to optimize action policies within a small number of episodes. Recent studies have tried to overcome this limitation by adding memory systems and architectural biases to improve learning speed, such as in Episodic Reinforcement Learning. However, despite achieving incremental improvements, their performance is still not comparable to how humans learn behavioral policies. In this paper, we capitalize on the design principles of the Distributed Adaptive Control (DAC) theory of mind and brain to build a novel cognitive architecture (DAC-ML) that, by incorporating a hippocampus-inspired sequential memory system, can rapidly converge to effective action policies that maximize reward acquisition in a challenging foraging task.



\end{abstract}
\section{Introduction}
With the advent of Deep Reinforcement Learning (DRL), the last decade has been marked by several historical landmarks in the field of Artificial Intelligence (AI), in terms of both scientific and societal impact. Reaching human-level performance in historically considered uniquely-human intellectual domains (such as chess \cite{silver:chess}, Go \cite{silver:go}, and real-time strategy video games \cite{vinyals:starcraft}) has led to renewed claims about its resemblance to human cognition. Some researchers have even proposed that DRL could be a valid metaphor to understand the brain \cite{marblestone:deep}. However, criticisms to this view have pointed out to at least one great fundamental difference between human learning and deep learning approaches: sample efficiency \cite{marcus:deep}.


Sample efficiency refers to the number of data samples required for a learning system to attain a target level of performance (see \cite{botvinick:RL} for a more detailed discussion on the topic). Current DRL models are sample inefficient because they require large amounts of training episodes to achieve human-level performance. An example of this limitation is the research on emergent tool-use in a hide-and-seek multi-agent game; where DRL agents took up to 380 million episodes to learn some of the more sophisticated behaviors \cite{openai:hide}.

Recent research in the field has sought to find solutions to the sample-inefficiency problem by introducing algorithms that make use of a memory system to bootstrap policy learning, such as in Episodic Reinforcement Learning (ERL) \cite{ERL:control}. In ERL, stored episodes are retrieved from past interactions with the environment to support decision-making when the agent encounters similar states. However, states are treated as being independent of one another. Therefore, ERL cannot capitalize on the sequential nature of experience that gives rise to highly informative state-transition statistics.



In this paper, we present a novel cognitive architecture well suited for state-of-the-art Machine Learning (ML) environments, whereby policy learning and action selection are based on a sequential memory system inspired by the human hippocampus \cite{buzsaki:hippocampus}. In particular, building upon the architectural principles of the Distributed Adaptive Control (DAC) theory of mind and brain, we derive a biologically-inspired architecture (DAC-ML) that leverages the continuous nature of experience to bootstrap policy learning in a sample-efficient manner. We demonstrate the learning efficiency of DAC-ML in a challenging foraging task and show how the sequential feature of the memory system is key for achieving such fast learning dynamics.



\section{Cognitive architecture}
\begin{figure}[t]
\includegraphics[width=\textwidth]{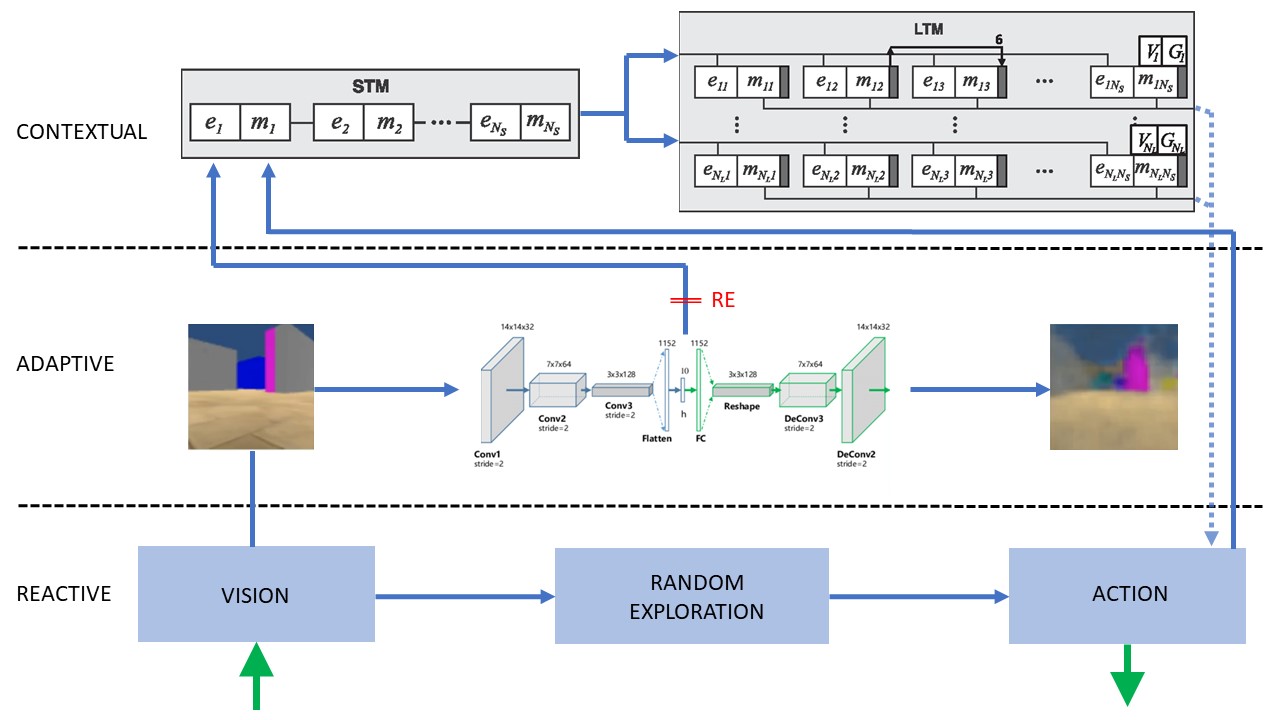}
\caption{Representation of the three layers of the DAC-ML architecture: Reactive (bottom), Adaptive (center), and Contextual (top). The Reactive layer implements a pseudo-random-walk generator that allows an efficient random exploration of the environment. The Adaptive layer comprises a convolutional autoencoder that generates compressed representations of the visual input to produce the state-action couplets to be conveyed to the STM buffer in the Contextual layer. The Contextual layer incorporates the STM and LTM buffers, along with an action selection algorithm, that allows the agent to perform policy learning and decision-making based on the stored episodic sequences. RE: reconstruction error; e: compressed state vectors; m: action; V: reward value; G: episodic sequence.}
\label{DAC-ML}
\end{figure}

DAC-ML is a biologically-inspired cognitive architecture composed of three layers (Reactive, Adaptive, and Contextual, see Fig. \ref{DAC-ML}). It is built based on the architectural principles of DAC of layered and hierarchical organization, that predicate that cognition is hierarchically distributed in different layers of control \cite{verschure:PTRS}, \cite{verschure:cogsci}, \cite{verschure:nature}. 

The Reactive layer represents the agent’s prewired reflexes and provides predefined responses that allow for simple but robust interaction with the environment. In DAC-ML, the Reactive layer incorporates a pseudo-random-walk algorithm that drives the agent to explore the environment and facilitates the input sampling initially required by the learning in the upper layers of the architecture (see Fig. \ref{DAC-ML}, bottom panel). 

The Adaptive Layer, where the perceptual learning takes place, contains a convolutional autoencoder that uses the sensory input sampled by the Reactive layer to build compressed representations of the state space (see Fig. \ref{DAC-ML}, middle panel). These state representations are then linked with the agent's action at every step to build the state-action couplets to be conveyed to the Contextual layer for policy learning. Moreover, the transmission of the couplets to the Contextual layer depends on the compressed representations surpassing a quality threshold, assessed by the convolutional autoencoder's corresponding reconstruction error.

The Contextual layer uses stored sequential episodic memories to perform policy learning and to guide real-time decision-making. It is composed of short-term memory (STM) and long-term memory (LTM) buffers, and an action selection algorithm (see Fig. \ref{DAC-ML}, top panel). The STM transiently stores the most recent sequence of state-action couplets generated by the agent's interaction with the environment, and it is updated following a first-in, first-out (FIFO) rule. Once a goal is encountered, the current sequence maintained in the STM is conveyed and stored in the LTM with its associated reward value. In the current implementation, the STM has a fixed length of 50 state-action couplets, while the capacity of LTM is limited to 100 sequences.

\subsubsection{Action selection with a sequential memory bias for rapid policy learning.}

The Contextual layer's action selection algorithm, while being invariant in itself, selects an action based on the recent history of observed states and performed actions. Concretely, at every time step, the observed compressed state is compared, based on a similarity metric (i.e., Euclidean distance), with all the state representations stored in LTM. The similarity scores are then weighted by an additional memory system storing the trigger values, which keep track of the recent history of selected state-action couplets (as in \cite{encarni:dac}). This procedure increases the eligibility of the couplets that come immediately after the recently selected ones. Hence, the resulting eligibility scores (i.e., similarity scores weighted by the trigger values) effectively bias the action selection towards continuing on recently visited sequences in LTM. 

Furthermore, a set of state-action couplets in LTM is selected based on their eligibility scores surpassing both absolute and proportional thresholds. This procedure enforces a winner-takes-all mechanism that has been shown to operate in the hippocampus \cite{lisman:emax}. The selected set of state-action couplets are then assigned the reward values associated with their corresponding sequences, weighted by an exponential decay based on each couplet's normalized distance to the end of the sequence. This mechanism favors the selection of actions that are closer in time to bigger rewards. Moreover, the reward values assigned to each couplet are normalized to the maximum reward associated with the selected couplets, thus implementing a relative reward valuing function. Lastly, the final action to be performed is randomly drawn from the probability distribution generated over the discrete action space by normalizing the sums of all the relative reward values associated with the actions of each selected couplet. Then, the trigger values are updated based on the selected couplets so that the couplets selection (and thus action selection) in the next time step is biased towards following the same sequences. This biasing mechanism is further reinforced by behavioral feedback \cite{verschure:nature}, whereby the input sampling caused by the selected action favors the storage of similar sequences of state-action couplets. At the same time, our implementation of the action selection as a probability distribution over the discrete action space keeps the Bayesian optimality feature of previous DAC models operating in a continuous action space \cite{verschure:cogsci}.

\section{Results}

\begin{figure}[ht]
\includegraphics[width=\textwidth]{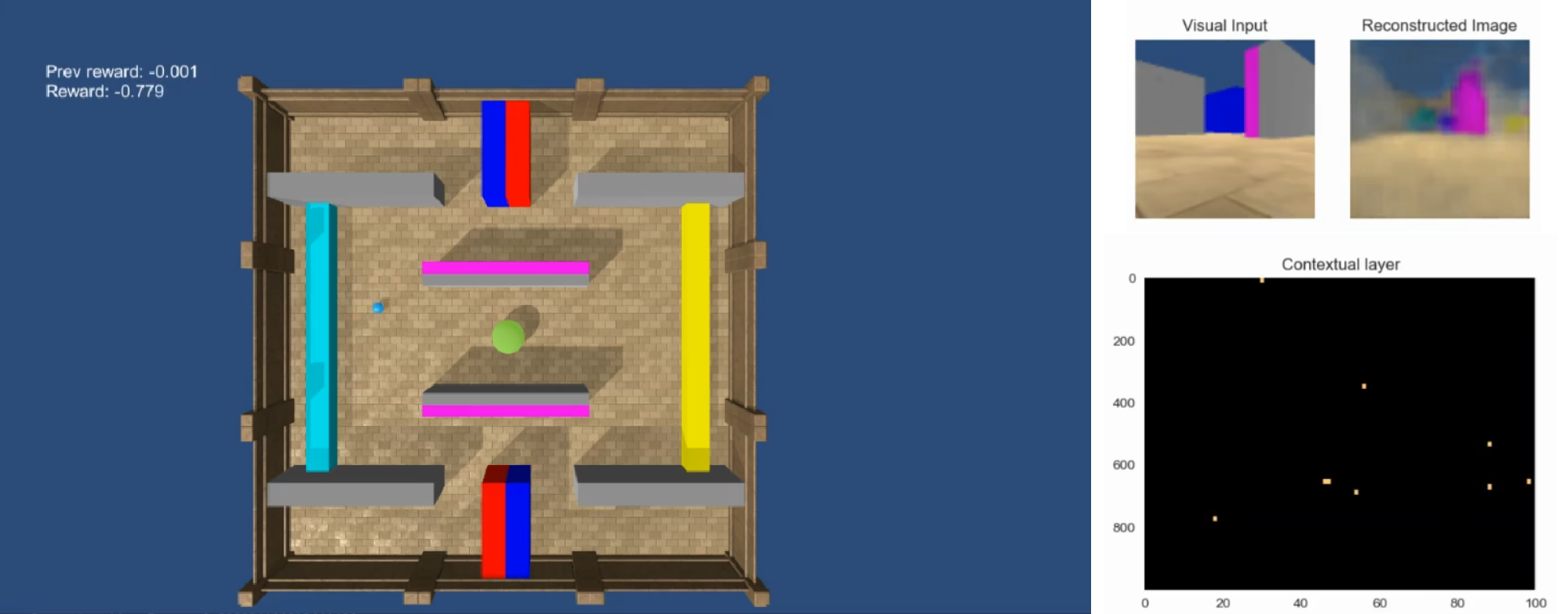}
\caption{Experimental setup. Left panel: a top-view of the foraging task modeled in Animal-AI. The small blue sphere represents the agent. The big green sphere at the center represents the reward. Top right panels: first-person view of the agent (left) and the reconstructed image generated by the Adaptive layer (right). Bottom right panel: representation of the selected couplets of the LTM in the Contextual layer.}
\label{expsetup}
\end{figure} 

We used the Animal-AI framework \cite{animal:ai} to create the simulation environment for testing the capabilities of the DAC-ML cognitive architecture. Animal-AI is an AI experimentation and evaluation platform specially designed to allow for cognitive testing in reinforcement learning and spatial navigation tasks. This framework imposes several realistic constraints regarding the type of input an agent can receive since the agents have to learn to navigate the 3D environment only guided by first-person observations in the form of raw pixel data.

To test the efficacy of DAC-ML in obtaining rewards by acquiring stable behavioral sequences, we tested it in a foraging task that requires the sequential encoding and retrieval of adequate visual cues in order to reach the reward location in an optimal amount of time. The environment is a colored maze (see Fig. \ref{expsetup}) with only one reward source in the center. The walls are painted with different colors to serve as visual cues for the agent. At the beginning of each episode, the agent is randomly allocated in one of the four corners of the maze and has to reach the center to acquire the reward. The reward value is initially set to 3, and it decays linearly with time until the end of the episode. An episode finishes either when the agent reaches the reward or when the time limit is reached.


 


\begin{figure}
\centering
\includegraphics[width=.75\textwidth]{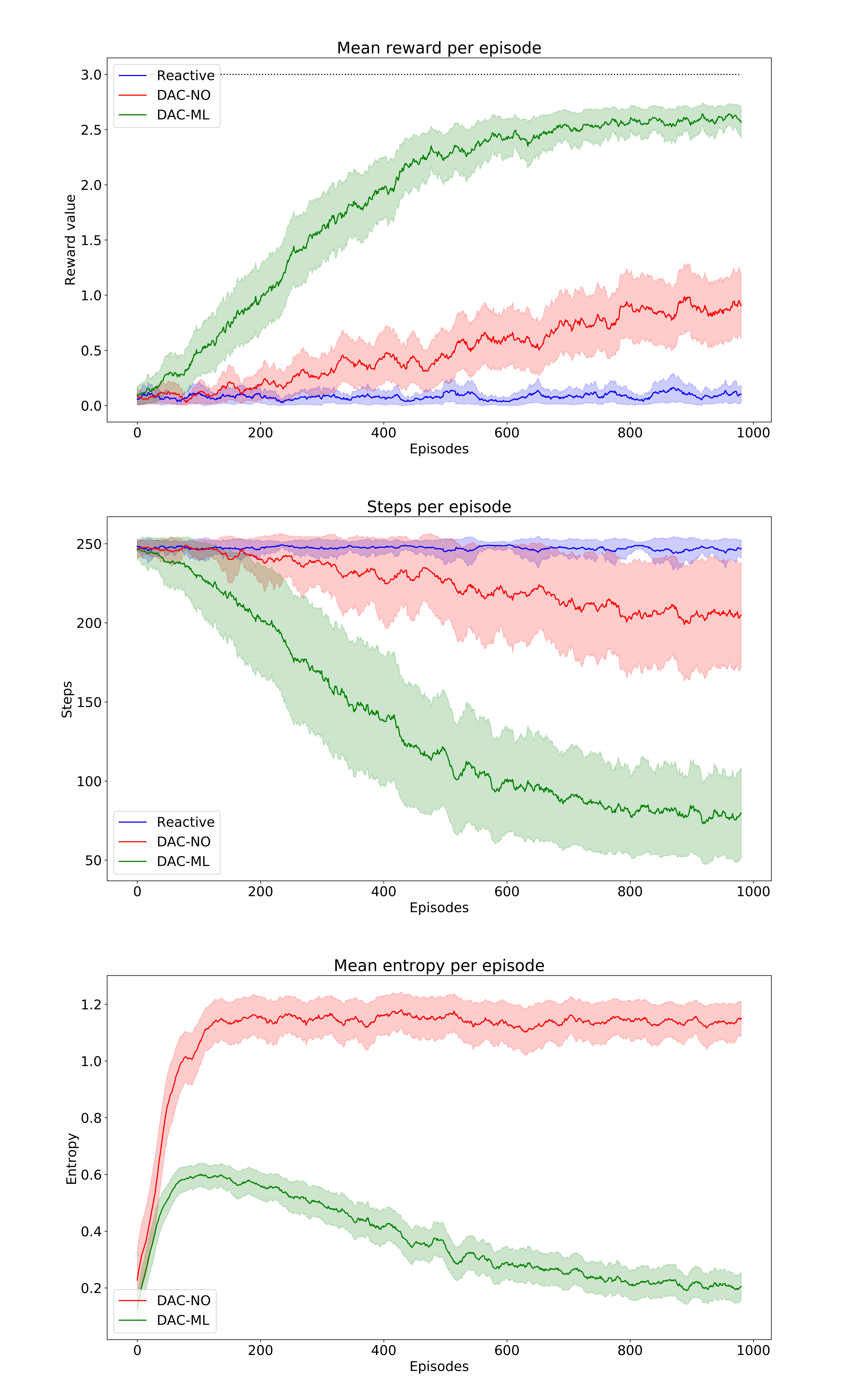}
\caption{DAC-ML rapidly stabilizes and maximizes reward acquisition in a foraging task. Top panel: mean reward per episode scores of DAC-ML (green), DAC-ML with no sequencing bias (red), and a reactive agent (blue). Center panel: mean steps per episode of the same agents. Bottom panel: mean entropy per episode for both DAC-ML agents. Entropy is computed over the probability distribution of actions at every step so that lower values indicate policy convergence and stability. All metrics have been computed by an averaged sliding window of 20 episodes over time.}
\label{results}
\end{figure}

We ran 20 simulations of 1.000 episodes each and compared the results of DAC-ML against two control agents: (1) a Reactive agent that randomly explores the environment, and (2) a version of DAC-ML with the same set of parameters but without incorporating the sequential memory bias, and thus treating the state-action couplets as independent units. The results show the capacity of DAC-ML to efficiently acquire rewards in the foraging task in relatively few episodes, as compared with the control agents. Fig. \ref{results}-top shows the difference in reward acquisition between the three agents. We show that DAC-ML is able to maximize the reward acquisition in less than 1.000 episodes. In addition, Fig. \ref{results}-center shows the number of steps (time) taken by the agents in every episode to reach the reward. These results demonstrate that DAC-ML can optimize the time it requires to reach the reward and that it generates rapid learning dynamics compared to a more standard cognitive architecture that does not incorporate a memory system based on episodic sequences. Finally, we also estimated the stability over time of the action selection (i.e., policy learning) by computing the mean entropy across the probability distributions over the action space throughout each episode. In the bottom panel of Fig. \ref{results}, we observe how the policy entropy also steadily decreases over time for DAC-ML (but not for the control agent), indicating convergence towards a stable action policy. Overall, these results demonstrate the advantage of incorporating a sequential memory bias in achieving rapid and stable policy learning in a cognitive architecture optimized by reinforcement learning.


\section{Discussion and Future Work}
In this paper, we have presented DAC-ML, a novel cognitive architecture based on the organizational principles of the Distributed Adaptive Control theory, that can efficiently maximize reward acquisition in a challenging foraging task. DAC-ML is a memory-guided decision-making architecture inspired by the hippocampus, that stores complete episodic sequences of state-action couplets to guide action selection. We showed that this biologically-inspired architectural bias promotes the rapid generation of stable and efficient goal-directed behavior, thus affording sample-efficient policy learning.

In contrast with classical RL where the policy is directly updated by the learning algorithm, in DAC-ML, policy learning is achieved through the interaction of the different components of its layered architecture. Since action-selection in DAC-ML follows a fixed rule, the main driver of policy learning is the constant acquisition and retrieval of new sequential memories. In other words, the behavioral policy of a DAC-ML agent is implicitly updated by the memories it forms in its interaction with the environment. Moreover, DAC-ML emphasizes the learning of complete behavioral sequences, whereas ERL usually treats its memory units as separate disconnected events. The incorporation of this hippocampus-inspired architectural bias bootstraps the formation of a behavioral feedback loop \cite{verschure:nature}, allowing DAC-ML to rapidly transition from an initial exploration phase to a more exploitative phase in which the behavior of the agent stabilizes over time while maximizing reward acquisition.  


With this work, we seek to contribute to the available proposed solutions to the sample-inefficiency problem in reinforcement learning by taking inspiration from empirical and theoretical research from the cognitive sciences. In future work, we plan to extend the results presented in this paper by directly comparing DAC-ML with state-of-the-art ERL models, and test it in multi-agent environments, such as in game-theoretic dilemmas \cite{freire:conventions} \cite{freire:tom} \cite{freire:multiagent}.


\section{Acknowledgements}
This research received funding from the H2020-EU project HR-Recycler, ID: 820742.
%
%

\end{document}